\documentclass[a4paper, 10pt, conference]{ieeeconf}      
\usepackage{FG2024}

\FGfinalcopy 

\IEEEoverridecommandlockouts                              
\usepackage{cite}
\usepackage{amsfonts}
\usepackage{algorithmic}
\usepackage{graphicx}
\usepackage{textcomp}
\usepackage{xcolor}
\usepackage{comment}
\usepackage{makecell,multirow,tabularx}
\usepackage{booktabs}
\usepackage{colortbl}
\usepackage{times}
\usepackage{epsfig}
\usepackage{subcaption}
\def\BibTeX{{\rm B\kern-.05em{\sc i\kern-.025em b}\kern-.08em
   T\kern-.1667em\lower.7ex\hbox{E}\kern-.125emX}}

\definecolor{Gray}{gray}{0.9}
\definecolor{LGray}{gray}{0.8}
\definecolor{LightCyan}{rgb}{0.01,0.01,0.35}

\newcolumntype{a}{>{\columncolor{Gray}}c}
\newcolumntype{b}{>{\columncolor{LGray}}c}

\usepackage{float}
\usepackage[hyphens]{url}
\usepackage{hyperref}
\usepackage{balance}
\usepackage{fancyhdr}

\begin{document}

\title{\textbf{Towards Inclusive Face Recognition Through Synthetic Ethnicity Alteration}
{\footnotesize
\thanks{This work was supported by the European Union’s Horizon 2020 Research and Innovation Program under Grant 883356}}
}

\author{Praveen Kumar Chandaliya$^\dag$ $^\ddag$, Kiran Raja$^\ddag$, Raghavendra Ramachandra$^\ddag$, Zahid Akhtar$^*$, Christoph Busch$^\ddag$\\
$^\dag$Sardar Vallabhbhai National Institute of Technology, Surat, India\\
$^\ddag$Norwegian University of Science and Technology, Norway\\
$^*$State University of New York Polytechnic Institute, Utica, New York, USA\\
pkc@aid.svnit.ac.in; akhtarz@sunypoly.edu; \\
\{kiran.raja, raghavendra.ramachandra,christoph.busch\}@ntnu.no
}

\maketitle

\thispagestyle{fancy}
\renewcommand{\headrulewidth}{0pt}
\fancyhf{}
\fancyhead[C]{2024 18th International Conference on Automatic Face and Gesture Recognition (FG)}





\fancyfoot[L]{979-8-3503-9494-8/24/\$31.00 \copyright 2024 IEEE}

\begin{abstract}
Numerous studies have shown that existing Face Recognition Systems (FRS), including commercial ones, often exhibit biases toward certain ethnicities due to under-represented data. In this work, we explore ethnicity alteration and skin tone modification using synthetic face image generation methods to increase the diversity of datasets. We conduct a detailed analysis by first constructing a balanced face image dataset representing three ethnicities: Asian, Black, and Indian. We then make use of existing Generative Adversarial Network-based (GAN) image-to-image translation and manifold learning models to alter the ethnicity from one to another. A systematic analysis is further conducted to assess the suitability of such datasets for FRS by studying the realistic skin-tone representation using Individual Typology Angle (ITA). Further, we also analyze the quality characteristics using existing Face image quality assessment (FIQA) approaches. We then provide a holistic FRS performance analysis using four different systems.  
Our findings pave the way for future research works in (i) developing both specific ethnicity and general (any to any) ethnicity alteration models, (ii) expanding such approaches to create databases with diverse skin tones, (iii) creating datasets representing various ethnicities which further can help in mitigating bias while addressing privacy concerns.

\end{abstract}


\section{Introduction}\label{sec:introduction}

Automated face recognition has made significant strides due to the rapid advancement of deep learning algorithms. The current state-of-the-art FRS \cite{PFE2019,Boutros_2022_CVPR} exhibits exceptionally high accuracy across different face recognition benchmarks in unconstrained environments (e.g., LFW \cite{LFWTech}, IJB-A \cite{IJBA2015}, AgeDB \cite{agedb}, ICD \cite{ICD2022}, and Mega Asia \cite{SSRNet2018}), with performance almost reaching near-ideal levels.  Several benchmarks focus on specific challenges, such as pose and aging \cite{Longitudinal2023}, which lead to substantial performance drops on other facial covariates. Despite improved performance, various studies have however questioned if face recognition systems perform equally well across various demographic groups \cite{DeBiasing2020}. Many FRS have exhibited lower accuracy rates for certain demographic groups with established empirical evidence leading to strong critique on biased performance \cite{DeBiasing2020} with commercial systems not being an exception\footnote{\url{https://news.mit.edu/2018/study-finds-gender-skin-type-bias-artificial-intelligence-systems-0212}}. Not different from FRS is the recognition phenomenon in humans who tend identify/recognize people of one's own race, often referred to as the other-race effect \cite{Klare2012, Demographics2023}.

\begin{figure}
		\centering
		\includegraphics[width=0.495\textwidth]{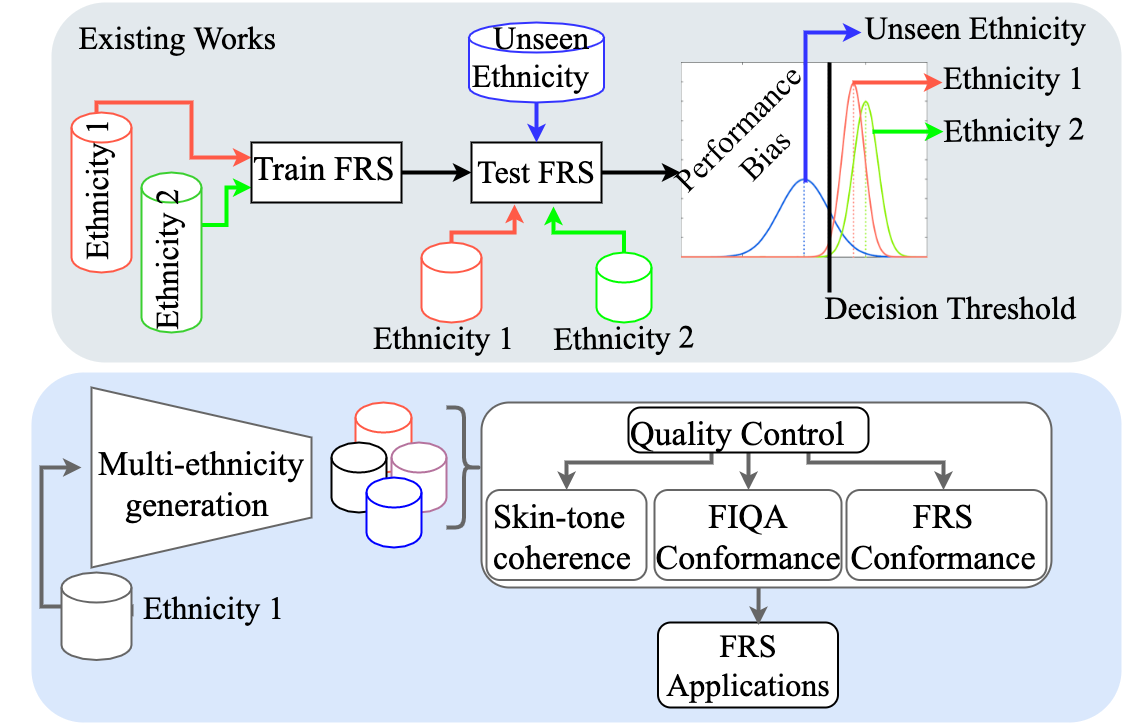}
		\caption{Our framework for generating multi-ethnicity data for biased performance for different ethnicities as compared to existing works in FRS.}
		\label{fig:framework-proposed}
\end{figure}

Among many demographic factors are the ethnicity and skin tone across diverse global populations noted to contribute significantly in biased performance. Lack of balanced large scale databases with multiple ethnicities and skin tone for training FRS plays a central role in making face recognition frameworks consistent across different ethnicities and skin tones. Prior studies on demographic bias primarily relied on real face image datasets such as RFW \cite{Wang_2019_ICCV}, YFCC \cite{YFCC100M}, UTKFace \cite{SCDAE2020}, MAAD-Face \cite{MAADFace2021}, BUPT-GlobalFace \cite{BUPTDataset}, and BUPT-Balancedface \cite{BUPTDataset}. In prior research, as depicted in Fig~\ref{fig:framework-proposed}, it has been common practice to train FRS on individuals belonging to one ethnic group and subsequently test them on individuals from a completely different ethnic group, such as training on Indian identities and testing on Black identities.

\begin{figure}[htp]
    \centering
    \includegraphics[width=0.48\textwidth]{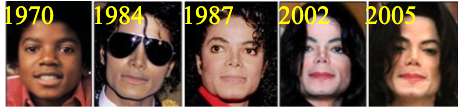}
    \caption{Michael Jackson's facial appearance change due to cosmetic surgeries and skin disorder disease resulting in changes in ethnical appearance.}
    \label{fig:Michael}
\end{figure}

Synthetic data has emerged as a promising alternative to real-world data, offering greater control over demographics and facial attributes manipulation. This allows for alignment with the desired demographic distributions while maintaining consistent verification and face image quality across diverse demographic groups \cite{melzi2024synthetic}. Moreover, synthetic face generators possess the potential not only to produce virtually infinite data, but also to mitigate privacy concerns that have prompted the discontinuation of established datasets and the enactment of regulatory frameworks such as the EU-GDPR \cite{GDPR}. We motivate our work taking the advantage of synthetic data generation to create identity of same person across different ethnicity to study the suitability for diversifying the datasets for training FRS. Further, some individuals may exhibit features of different ethnicities before and after due to extensive cosmetic surgeries \cite{Jackson93} or skin disorder disease \cite{Jackson2012} as shown in Fig. \ref{fig:Michael}. A significant transformation in facial features around the nose and skin tone can be noticed from Fig. \ref{fig:Michael}. We thus assert transforming the ethnicity of a person across different ethnicities can help in diversifying and increasing the datasets for training FRS as shown in Fig~\ref{fig:framework-proposed}. To validate our assertion, we pose three research questions:  Q1. Does synthetic ethnicity alteration and skin tone modification using Generative Adversarial Networks (GANs) find use in FRS for representing a subject in different ethnicities? Q2. Does synthetic ethnicity alteration and skin tone modification of a subject's facial image  resemble ITA of real faces? Q3. Do FIQA and FRS measures for synthetic ethnicity-altered image datasets indicate dependable performance, similar to real datasets of different ethnicities?

In the course of answering these questions, we first construct a balanced dataset of multiple ethnicities (i.e., Indian, Asian and Black) and make use of generating transformed ethnicities. We specifically make use of existing image-to-image translation models and manifold learning based translation models to generate different ethnicity representation of any subject. We generate a new synthetic dataset from each of these settings resulting in $45,000$ images representing three ethnicities. We then systematically assess the skin-tone of transformed ethnicity to verify the representative nature of skin-tone of real data using ITA. We then assess if the images provide performances similar to real data by analysing the FIQA and subsequently FRS. For the analysis using FIQA, we provide Error-vs-Discard Characteristic curves (EDC) using three different FRS. Complementary to that, we use another fourth FRS to benchmark the verification performance in a cross-ethnicity scenario (e.g., a subject from Indian ethnicity represented in Asian). Our contributions in this work include:
\begin{itemize}
    \item We present a novel framework for generating ethnicity-altered synthetic face images for given subjects utilizing image-to-image translation models and manifold learning. To the best of our knowledge, this is the first work investigating synthetic ethnicity alteration and skin tone modification without relying on pre-trained models. 
    \item A new racially balanced dataset comprising $45,000$ images balanced across three different ethnicity (Asian, Black, and Indian) is introduced to facilitate the study of ethnicity alteration dynamics. 
    \item  We introduce Individual Typology Angle (ITA) as a new metric for measuring skin-tone transformation in synthetic data generation approaches for measuring representative skin-tone of ethnicity.
    \item A comprehensive cross-ethnicity face verification analysis is presented based on four distinct face recognition systems: FaceNet \cite{FaceNet2015}, PFE \cite{PFE2019} and two variants of ElasticFace 
    \cite{Boutros_2022_CVPR}. The analysis is complemented with a detailed analysis of FIQA using three different FRS. 		
\end{itemize}

The work while deepening our understanding of cross-ethnicity face recognition accuracy, skin tone analysis, and FIQA on synthetically generated face images with ethnicity-altered effects, can further help in enhancing fairness and equity, mitigating bias, improving generalization, addressing security and privacy concerns, and supporting future research. We aim to promote fairness and accuracy in FRS technology while addressing biases and ensuring inclusivity across diverse populations through synthetic data generation. The rest of the paper is organized as follows: Section~\ref{sec:relatedwork} reviews related works, Section~\ref{sec:investigation} explains investigation methodology, Section~\ref{sec:dataset} discusses the details of the newly created racial balance and publicly available test dataset, Section~\ref{sec:qualitativeevalation} presents a qualitative study on ethnicity dataset of Asian, Black, and Indian subjects. Section~\ref{sec:quantitative} further presents a quantitative analysis using FRS and ITA, and Section~\ref{sec:challenges} presents unaddressed challenges along with a discussion in Section~\ref{sec:discussion} and Section \ref{sec:conclusion} concludes the paper.

\section{RELATED WORK}\label{sec:relatedwork}
We present a brief of existing works in face recognition and image-to-image translation, as these topics are pertinent to the subject matter addressed in this study.

\subsection{Demographic differentials in face recognition}
    Automatic facial recognition systems are prone to bias. The Cross-Race Effect is a bias, characterized by improved accuracy in recognizing faces of the same race as the system's training data compared to faces of a different race \cite{Steven2012}. Klare et al. \cite{JainDemographic2012} showed that the performance of face recognition systems is strongly influenced by demographics attribute. 
    The National Institute of Standards and Technology (NIST) has shown that state-of-the-art algorithms continue to display a higher rate of false positives among individuals from West and East Africa and East Asia, whereas those of Eastern European descent have the lowest false positive rate \cite{FRVT2019}.
	Wang et al. \cite{Wang_2019_ICCV} demonstrated that ``All algorithms and APIs perform the best on Caucasian testing subsets, followed by Indian, and the worst on Asian and African. This is because the learned representations predominantly trained on Caucasians will discard useful information for discerning non-Caucasian faces.''
	This work concentrates on skin tone and ethnicity bias by constructing a balanced ethnicity dataset. The newly created in-house Ethnicity Alteration Training (ETAT) Dataset and synthetically generated cross-ethnicity images are used to understand the performance of FRS in our work.  

	\subsection{Image-to-Image Translation in Face Alteration}
	Face alteration can be viewed as a subproblem within image-to-image translation, which involves converting an input image from one domain to another while retaining its original identity. 
    Unsupervised image-to-image translation using GANs has attracted increasing interest in facial attribute editing. CycleGANs \cite{cyclegan} have recently proven successful in two-domain image-to-image translation by employing a cycle consistency loss function, which diminishes the necessity for a paired dataset (unlike \cite{isola2018imagetoimage}) and preserves crucial attributes between the input and translated images. 
	
	StarGAN \cite{StarGAN2018} proposed a single conditional generator to learn mapping among multiple domains, with its discriminator functioning as an auxiliary classifier to classify attributes within these domains. Manifold learning-based FGAN \cite{fgan} was explicitly introduced for racial transformation to enhance training stability compared to traditional image-to-image translation approaches. In our work, we use these models (CycleGAN, StarGAN and FGAN) for generating ethnicity altered images. Specifically, we use one generator with one encoder and multiple decoders, each corresponding to a distinct ethnicity.
 
 %
	
	\begin{figure}
		\centering
		\includegraphics[width=7.5cm]{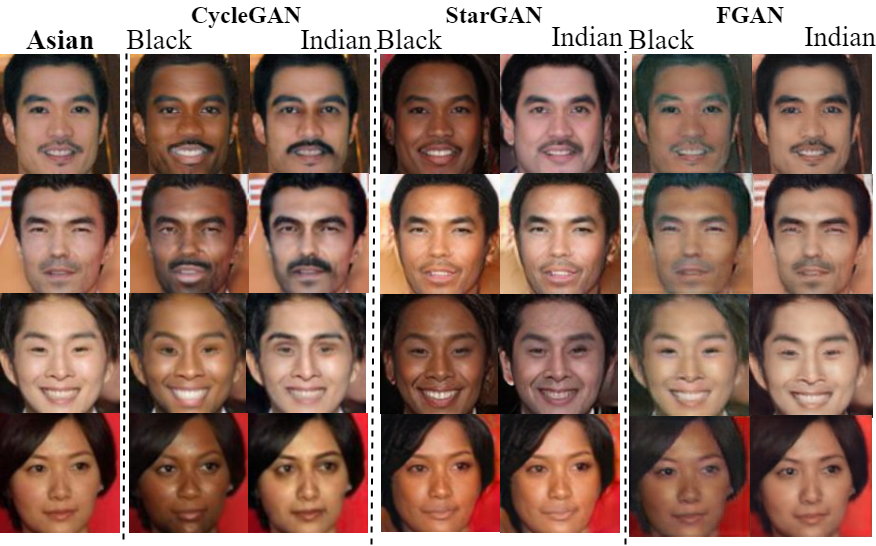}
		\caption{Ethnicity alteration results on Asian faces.}
		\label{fig:treeganasian}
		\vspace*{-0.3cm}
	\end{figure}
	
	\section{Investigation Methodology}\label{sec:investigation}
    The primary objective of this work was to study the feasibility of altering the ethnicity synthetically for FRS applications. To accomplish this, a well-structured and balanced dataset which we refer to as Ethnicity Alteration Training (ETAT) dataset (detailed in Section \ref{sec:dataset}) was created representing three distinct ethnic groups (Asian, Black, and Indian). The GAN-based models were trained to alter the ethnicities using six different combinations: Asian $\mapsto$ Black, Asian $\mapsto$ Indian, Black $\mapsto$ Asian , Black $\mapsto$ Indian , Indian $\mapsto$ Asian, and Indian $\mapsto$ Black. The trained models were applied to the RFW \cite{Wang_2019_ICCV} dataset to perform the ethnicity transformations. We first analyze the skin-tones of generated images. Further to that, FIQA using Error-vs-Discard characteristic (EDC) \footnote{\url{https://www.iso.org/standard/79519.html}} analysis is conducted to measure the suitability of ethnicity-altered images. To assess the effects of ethnicity alterations on FRS, four pre-trained deep FRS models (FaceNet, PFE, Elastic Arc+, and Elastic Cos+) were studied.  The results were measured using three complementary metrics: (i) cross-ethnicity skin tone transformation utilizing an ITA, (ii) FIQA analysis using EDC and (iii) cross-ethnicity face verification performance using FNMR and FMR.

	\section{Dataset}\label{sec:dataset}
	We utilized the Ethnicity Alteration Training (ETAT) dataset exclusively for training and the RFW dataset \cite{Wang_2019_ICCV} for testing. The primary reason for this choice was the balanced ethnic representation in the ETAT dataset, which was lacking in the RFW dataset. Nonetheless, the RFW dataset offered an advantage regarding multiple samples per subject, which were leveraged for enhanced analysis during testing. 
	
	\subsection{Ethnicity Alteration Training (ETAT) Dataset}\label{sec:abirbd}
	
	Existing large-scale face image datasets, such as CACD \cite{CACD2014}, FFHQ \cite{StyleGAN2019}, Celeba \cite{celeba}, AgeDB \cite{agedb}, MORPH \cite{MORPH2006}, CASIA \cite{Casia2014}, and Maga Asia \cite{SSRNet2018}, are predominantly composed of Caucasian faces and significantly under-represent other ethnicities, particularly East Indian skin tone. 
    To address these limitations, a more balanced dataset is introduced which contains 45,000 images evenly distributed across Asian, Black, and Indian races. We further refer to this dataset as the "Ethnicity Alteration Training (ETAT) Dataset".

	The process of developing the ETAT dataset involved initially predicting the race of images within existing databases and then adding images of Indian ethnicity through dedicated web crawling efforts\footnote{\url{https://github.com/deepanprabhu/duckduckgo-images-api}}. This subset of web-crawled images constitutes a comprehensive Indian dataset from various states 
	 containing approximately 57,000 images exclusively from Indians. For ground truth annotation in publicly available datasets (e.g., CelebA \cite{celeba}, FFHQ \cite{StyleGAN2019}, AgeDB \cite{agedb}, Mega Asia \cite{huang2016unsupervised}, and CASIA-WebFace \cite{Yi2014LearningFR}), state-of-the-art methods (e.g., DeepFace \cite{serengil2021lightface}) were utilized to predict ethnicity and age in these datasets. All images were processed using an advanced face detection technique (i.e., MTCNN \cite{mtcnn}), including cropping, alignment, and resizing ($128 \times 128$), to ensure consistency and quality. After filtering out blurry, non-readable, and low-quality images, $15,000$ high-quality images representing Asian, Black, and Indian ethnicities were selected for training. This is the first Indian subset to represent an Indian dataset from different states (details are provided in supplementary material). The ETAT dataset fills a crucial gap in Indian face representation in the FRS and facilitates research on mitigating race bias for more inclusive AI systems.


	\begin{figure}
		\centering
		\includegraphics[width=7.5cm]{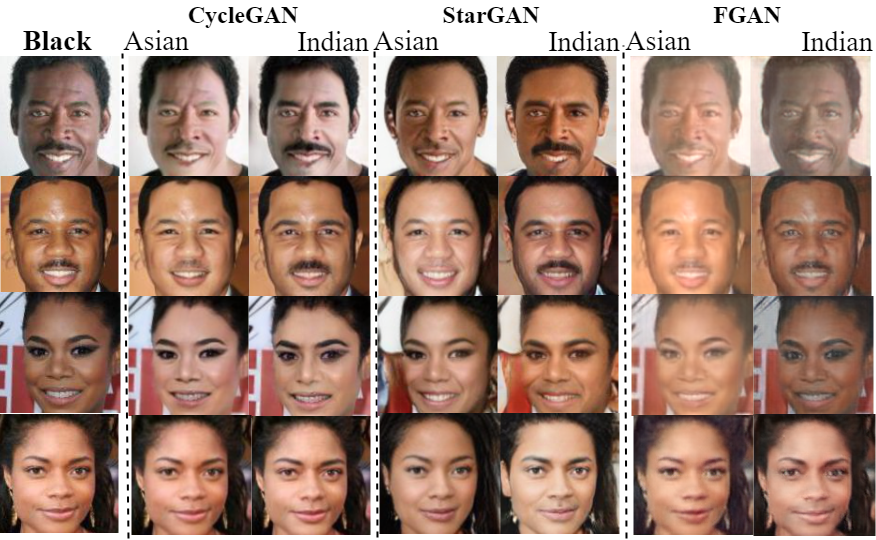}
		\caption{Ethnicity alteration results on Black faces.}
		\label{fig:treeganblack}
		\vspace*{-0.3cm}
	\end{figure}

	\subsection{Racial Faces in-the Wild}
	To evaluate the performance of the CycleGAN, StarGAN, and FGAN models for modifying ethnicity and skin tone, we utilized the Racial Faces in the Wild (RFW) test dataset for the testing phase \cite{Wang_2019_ICCV}. This dataset contains four testing subsets: Caucasian, Asian, Indian, and African/Black, with each subset comprising approximately 10K images of 3K individuals for face verification. To assess the quality of the test dataset, we employed the MagFace system, which estimates the face sample quality based on the magnitude of embeddings \cite{meng2021magface}. We removed $50\%$ of the images with the lowest magnitude, which typically contained low-quality images, in the test dataset. After this semi-automated process, our testing dataset comprised Asian (1240/3579 subjects/images), Black (1386/3513), and Indian (1277/3279) ethnicities.

	\section{Qualitative Evaluation}\label{sec:qualitativeevalation}
	In this section, we present a series of quantitative comparisons to provide readers with a comprehensive understanding of ethnicity alteration based on our findings.
	
	\subsection{Asian $\mapsto$ Black, Indian}
\begin{figure}
		\centering
		\includegraphics[width=8.5cm]{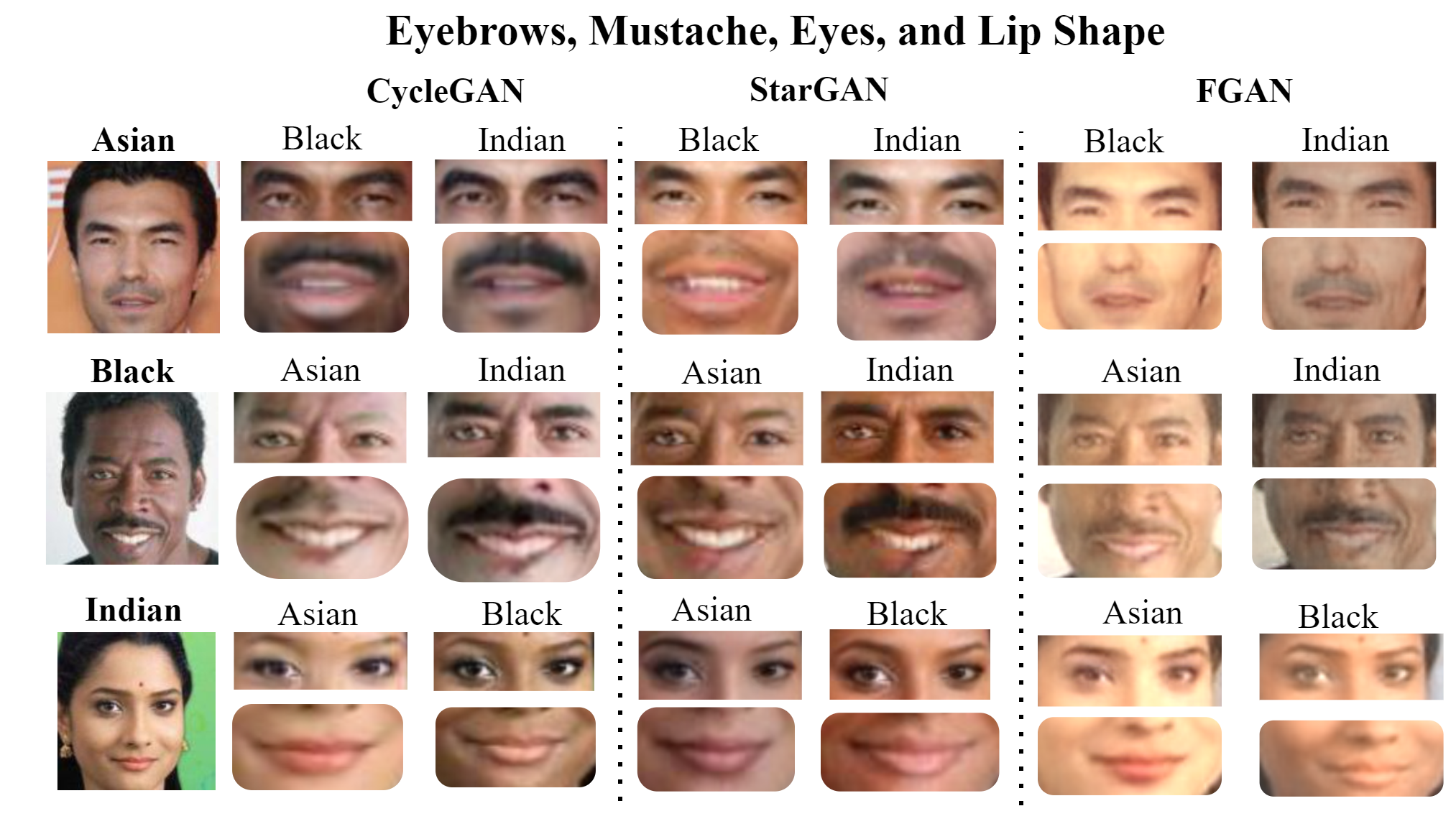}
		\caption{Illustration of ethnicity alteration effect on Asian, Black, and Indian with CycleGAN and StarGAN.}
		\label{fig:zoomfigure}
	\end{figure}

	Figure \ref{fig:treeganasian} illustrates ethnicity-altered images transitioning from Asian to Black and Indian ethnicity. We can observe that CycleGAN, StarGAN, and FGAN models execute structural changes in eyes, lip shape, and facial color characteristic of Black and Indian ethnicity (see Figure \ref{fig:zoomfigure}). All models generated realistic images with altered ethnicity from the original ethnicity. 
	
	\subsection{Black $\mapsto$ Asian, Indian}
	Figure \ref{fig:treeganblack} demonstrates that all models can effectively translate images from Black to Indian ethnicity. Indian ethnic groups, in contrast to Asians, commonly exhibit prominent eyelids, well-defined nasal bridges with tip projection, and relatively darker and more uneven skin tones as seen in Figure \ref{fig:treeganblack}.
	
	\begin{figure}
		\centering
		\includegraphics[width=7.4cm]{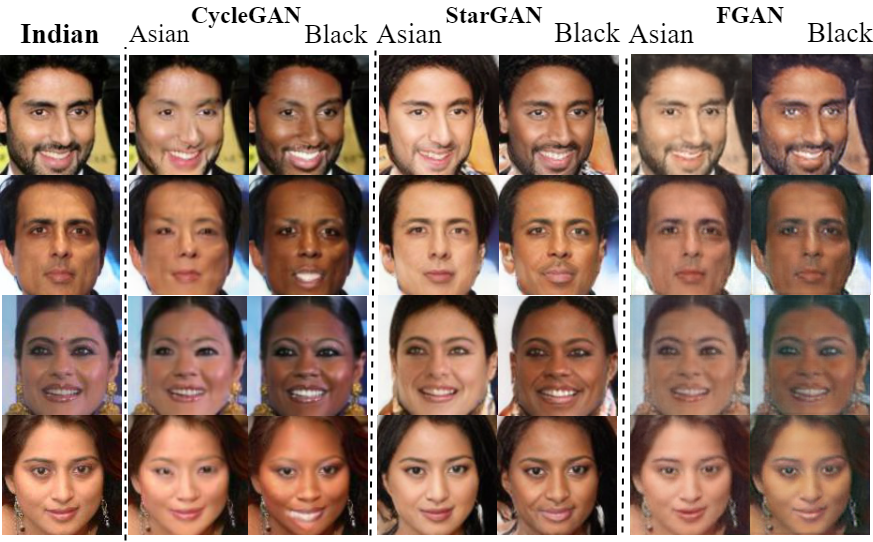}
		\caption{Ethnicity alteration results on Indian faces.}
		\label{fig:treeganindian}
	\end{figure}
	
	\subsection{Indian $\mapsto$ Asian, Black}
	We can observe that the eyes have narrowed, the skin tones have adjusted according to ethnicity in Fig~\ref{fig:treeganindian}. The eyelids, a well-defined nasal bridge with tip projection and relatively darker and more uneven skin tones when transitioning from Indian to Asian or Black ethnicity can further be observed. Additionally, Asian groups tend to exhibit wider intercanthal widths and smaller eye openings, which are notably distinct in the ethnicity-altered faces.

	\section{Quantitative Evaluation}\label{sec:quantitative}

	To assess the consequences of altering ethnicity on the Face Recognition System (FRS), we suggest three quantitative measures:
\begin{enumerate}
    \item \textbf{Skin tone estimation:} This measures the accuracy of estimating the skin tone from the original image to an altered image of a different ethnicity.
    \item \textbf{Face image quality assessment performance:} This assessment presents an in-depth analysis of the correlation between FRS and FIQA on an altered image of a different ethnicity \cite{PraveenISO}. To evaluate the face image quality assessment performance, we follow the methodology by Schlett et al. \cite{Torsten2022} using  Error versus discard characteristics (EDC).
    \item  \textbf{Cross-ethnicity face verification:} This involves comparing the original, unmodified image with the generated, altered image of the same person of a different ethnicity. This served as a test for identity persistence.
\end{enumerate}
These three criteria aim to provide a comprehensive evaluation of the impact of ethnic alteration on FRS.

\begin{figure}
    \centering
    \begin{subfigure}[b]{0.32\linewidth}
        \centering
        \caption*{\textbf{Asian}}		
         \includegraphics[width=\textwidth]{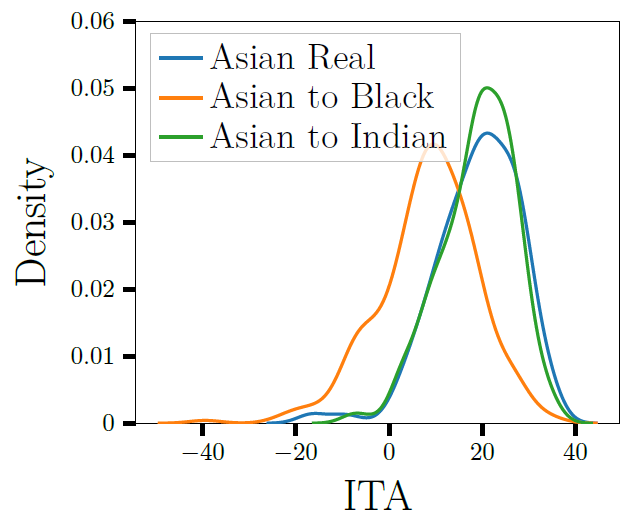}
        \label{fig:Cycleblackita}
    \end{subfigure}
    \hfill
    \begin{subfigure}[b]{0.32\linewidth}
        \centering
        \caption*{\textbf{Black}}
        \includegraphics[width=\textwidth]{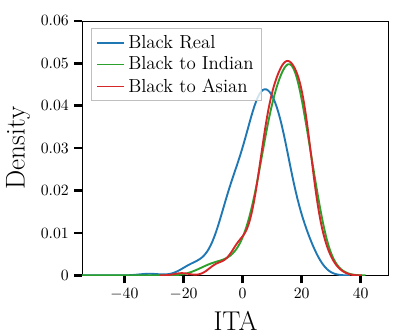}
        \label{fig:Cycleasianita}
    \end{subfigure}
    \hfill
    \begin{subfigure}[b]{0.32\linewidth}
        \centering
        \caption*{\textbf{Indian}}
        \includegraphics[width=\textwidth]{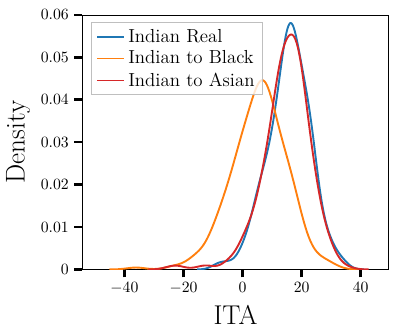}
        \label{fig:Cycleindianita}
    \end{subfigure}
    \centering
    \begin{subfigure}[b]{0.32\linewidth}
        \centering
        \includegraphics[width=\textwidth]{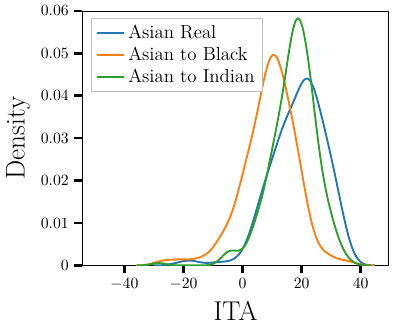}
        \label{fig:startganblackita}
    \end{subfigure}
    \hfill
    \begin{subfigure}[b]{0.32\linewidth}
        \centering
        \includegraphics[width=\textwidth]{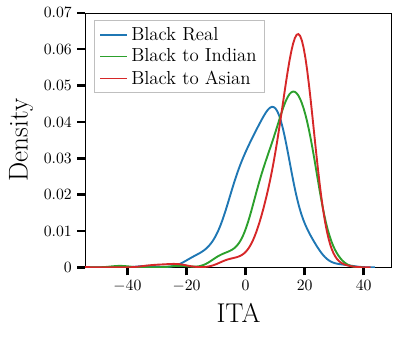}
        \label{fig:startganasianita}
    \end{subfigure}
    \hfill
    \begin{subfigure}[b]{0.32\linewidth}
        \centering        \includegraphics[width=\textwidth]{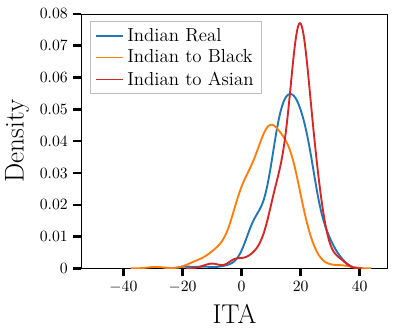}
        \label{fig:startganindianita}
    \end{subfigure}
    \centering
    \begin{subfigure}[b]{0.32\linewidth}
        \centering
        \includegraphics[width=\textwidth]{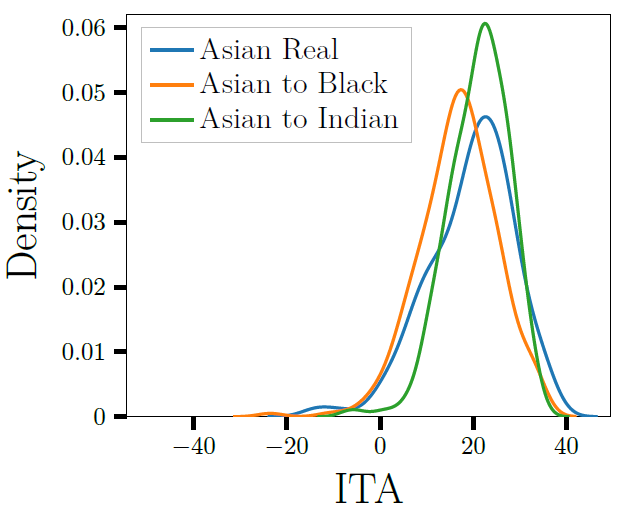}
        \label{fig:treeganblackita}
    \end{subfigure}
    \hfill
    \begin{subfigure}[b]{0.32\linewidth}
        \centering
        \includegraphics[width=\textwidth]{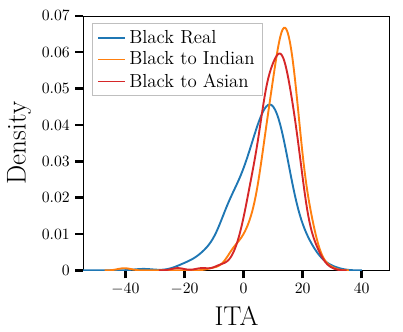}
        \label{fig:treeganasianita}
    \end{subfigure}
    \hfill
    \begin{subfigure}[b]{0.32\linewidth}
        \centering
        \includegraphics[width=\textwidth]{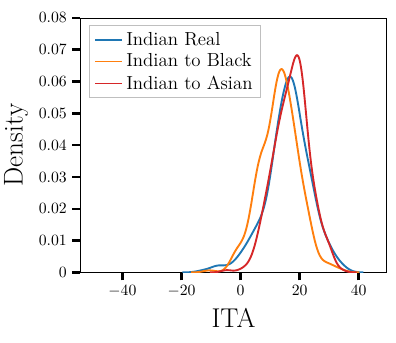}
        \label{fig:treeganindianita}
    \end{subfigure}
    \caption{Individual Typology Angle (ITA): Skin tone
        distribution of Asian, Black, and Indian races compared to  CycleGAN (row1), StarGAN (row2), and FGAN (row3).
    }
    \label{fig:skintoneita}
    \vspace{-.5cm}
\end{figure}

\subsection{Skin Tone Estimation}
To analyze if the ethnicity alteration represents the natural skin tone of the three ethnicities, we segmented facial regions using facial landmarks with Dlib and cropped the left and right cheek regions based on ISO/IEC 29794-5 \cite{ISO2023}. We randomly selected a crop region for analysis. Then, we calculate the Individual Typology Angle (ITA) \cite{ITA1991} using the equation  $ITA=[arctan(L*-50/b*]*180/\pi$, where L* represents luminance ranging from black (0) to white (100) and b* represents luminance ranging from yellow to blue \cite{ITA1991}. ITA skin color types are classified into six groups, ranging from Very Light to Dark skin: Very Light $(>55^{\circ})$, light  $(41^{\circ}  \to <55^{\circ})$, intermediate $(28^{\circ} \to  <41^{\circ})$, tan $(10^{\circ} \to  <28^{\circ})$, brown $(-30^{\circ} \to  <10^{\circ})$ and dark $(<-30^{\circ})$. Figure~\ref{fig:skintoneita} shows the distribution of skin tones in the test images corresponding to target-generated ethnicity images. It is evident that the StarGAN and FGAN ethnicity-altered images transformed the skin tone accordingly. Furthermore, the ITA remained relatively stable in the regular face images. Specifically, ITA classifies black-face images as intermediate to dark, Asian face images as light to intermediate, and Indian face images as tan to brown suggesting the skin tones to be consistent with real world data.

\begin{figure*}[htp]
		\centering
    \begin{subfigure}[b]{0.33\linewidth}
            \centering
            \caption*{\textbf{Asian}}
            \includegraphics[width=\textwidth]{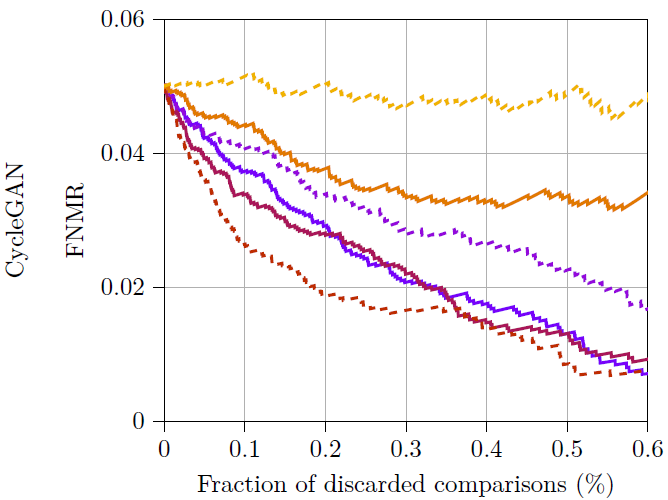}
    \end{subfigure}
    \begin{subfigure}[b]{0.31\linewidth}
        \centering
        \caption*{\textbf{Black}}
        \includegraphics[width=\textwidth]{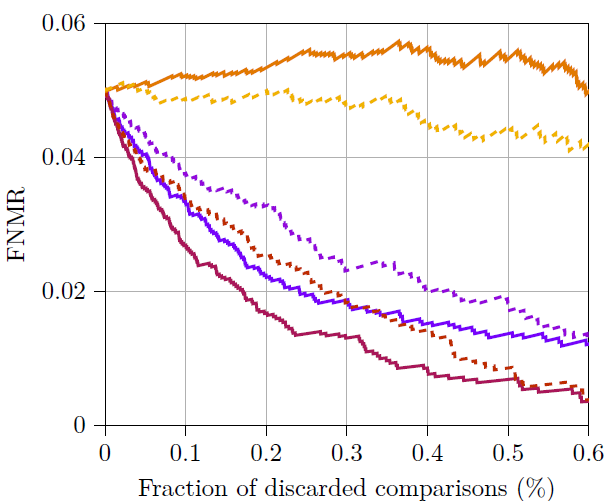}
    \end{subfigure}
    \begin{subfigure}[b]{0.32\linewidth}
        \centering
        \caption*{\textbf{Indian}}
        \includegraphics[width=\textwidth]{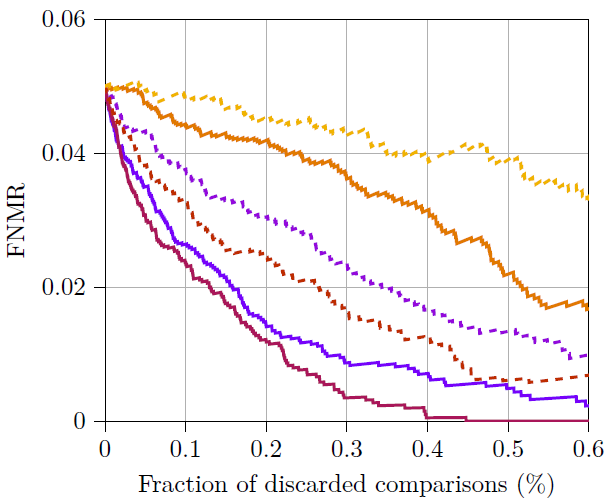}
    \end{subfigure}
   \begin{subfigure}[b]{0.335\linewidth}
        \centering
        \includegraphics[width=\textwidth]{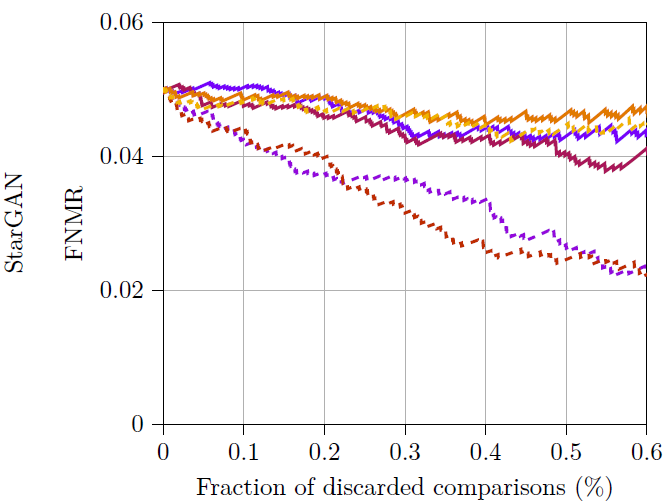}
        \label{fig:AsianCRFIQAL}
    \end{subfigure}
    \begin{subfigure}[b]{0.31\linewidth}
        \centering
        \includegraphics[width=\textwidth]{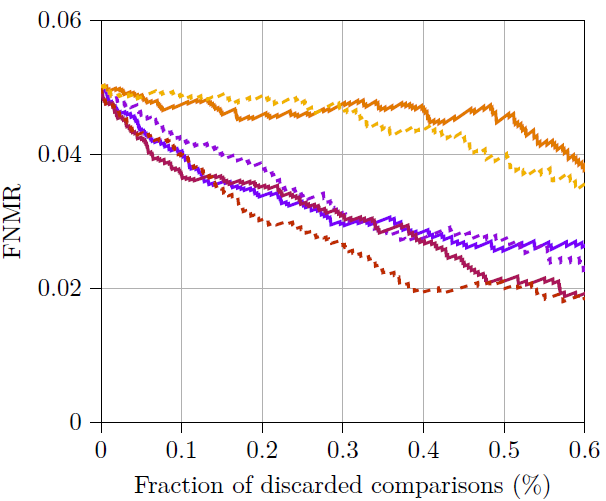}
    \end{subfigure}
    \begin{subfigure}[b]{0.32\linewidth}
        \centering
        \includegraphics[width=\textwidth]{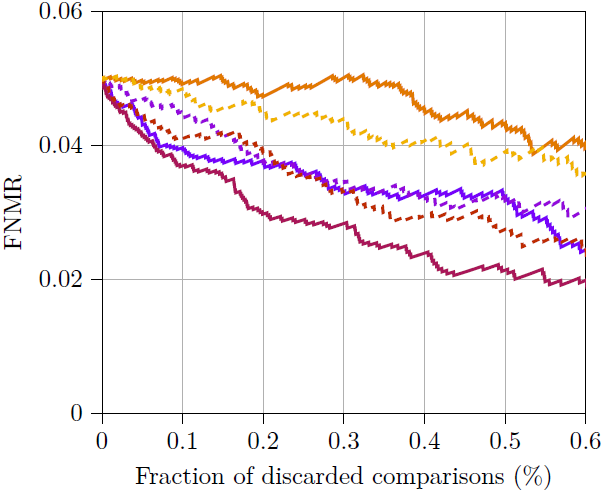}
    \end{subfigure}
        \centering
        \begin{subfigure}[b]{0.335\linewidth}
        \centering
        \includegraphics[width=\textwidth]{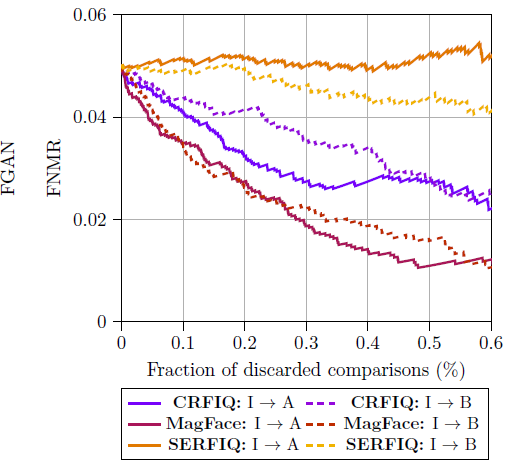}
    \end{subfigure}
    \begin{subfigure}[b]{0.31\linewidth}
        \centering
        \includegraphics[width=\textwidth]{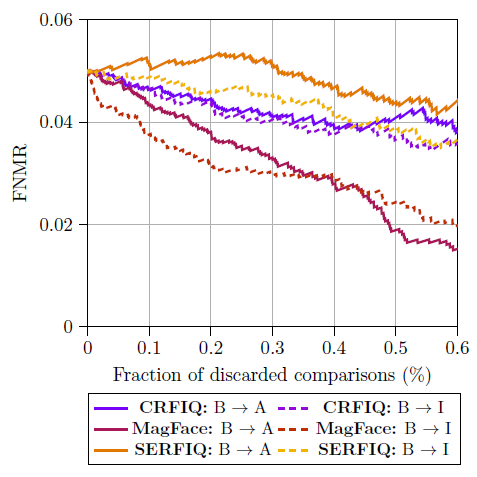}
    \end{subfigure}
    \begin{subfigure}[b]{0.32\linewidth}
        \centering
        \includegraphics[width=\textwidth]{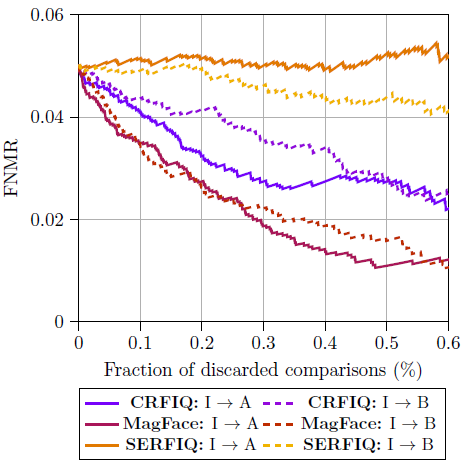}
    \end{subfigure}
    \caption{EDC plot: performance evaluation of ArcFace on CRFIQA, MagFace, and SERFIQ with CycleGAN (first row), StarGAN (second row), and FGAN (third row).}
    \vspace*{-0.3cm}
    \label{fig:edc}
\end{figure*}

\begin{table*}[htbp]
    \centering
    \Large
    \caption{Cross ethnicity face verification rate (\%) of face recognition systems on Asian (mated comparison:11,443, non-mated comparison:14305), Black (mated comparison:10,195, non-mated comparison:15,036), and Indian (mated comparison:10,143, non-mated comparison:13,097) ethnicity. The FNMR is evaluated at two FMR thresholds for four face recognition systems.}
    \resizebox{0.96\textwidth}{!}{
\begin{tabular}{lc|aaa|bbb}
\hline
& \multicolumn{7}{c}{FNMR@FMR=0.1\% / FMR=0.01\%}\\
\cline{3-8}
& Real & CyleGAN & StarGAN & FGAN & CyleGAN & StarGAN & FGAN \\
\cline{2-8}
FRS & Asian  &  \multicolumn{3}{c|}{Asian $\mapsto$ Black} &  \multicolumn{3}{c}{Asian $\mapsto$ Indian} \\
\hline
\hline
\textbf{FaceNet}   &      74.82  /   49.49 & 59.73  /   30.60 & 34.15  /   11.89 & 65.69  /   27.28 & 60.21  /   42.14 & 31.13  /   18.19 & 59.73  /   30.60\\
\textbf{PFE}     &        82.17  /   45.87 & 95.92  /   82.76 & 85.85  /   55.06 & 99.48  /   95.46 & 98.23  /   93.63 & 83.65  /   67.35 & 98.86  /   93.42\\
\textbf{Elastic Arc+}  &  99.94  /   99.77 & 97.81  /   88.84 & 90.78  /   57.04 & 99.92  /   99.42 & 97.81  /   88.84 & 91.83  /   84.34 & 99.78  /   98.54\\
\textbf{Elastic Cos+}  &  100.0  /   99.87 & 98.25  /   90.77 & 92.65  /   59.93 & 99.89  /   98.75 & 99.74  /   97.86 & 93.63  /   71.30 & 99.76  /   99.44\\ 
\hline 
\hline
& Black  &  \multicolumn{3}{c|}{Black $\mapsto$ Asian} &  \multicolumn{3}{c}{Black $\mapsto$ Indian} \\
\hline
\hline                     
\textbf{FaceNet}    &     82.17  /   45.87 & 49.30  /   23.16 & 28.54  /   12.66 & 75.59  /   60.86 & 42.37  /   17.57 & 29.76  /   15.85 & 69.95  /   45.88\\
\textbf{PFE}      &       99.93  /   99.51 & 99.49  /   95.48 & 87.39  /   57.81 & 99.77  /   98.99 & 95.33  /   85.70 & 82.17  /   64.34 & 99.37  /   97.61\\
\textbf{Elastic Arc+}  &  99.96  /   99.87 & 99.57  /   98.12 & 90.64  /   67.03 & 99.93  /   99.82 & 98.48  /   87.48 & 89.57  /   55.57 & 99.84  /   95.81\\
\textbf{Elastic Cos+}  &  99.93  /   99.80 & 97.04  /   90.83 & 88.58  /   75.13 & 99.92  /   99.86 & 98.46  /   93.15 & 91.06  /   81.21 & 99.80  /   98.64\\
\hline 
\hline
& Indian &  \multicolumn{3}{c|}{Indian $\mapsto$ Black} &  \multicolumn{3}{c}{Indian $\mapsto$ Asian} \\
\hline
\hline                     
\textbf{FaceNet}    &     89.39  /   57.44 & 60.92  /   41.05 & 51.56  /   24.50 & 85.33  /   68.79 & 67.64  /   52.46 & 60.07  /   28.06 & 85.74  /   51.80\\
\textbf{PFE}      &       99.86  /   99.42 & 96.32  /   91.04 & 94.60  /   87.41 & 99.82  /   99.02 & 99.91  /   99.79 & 94.44  /   81.03 & 99.74  /   97.57\\
\textbf{Elastic Arc+}  &  99.86  /   99.79 & 98.31  /   90.16 & 93.98  /   85.58 & 99.85  /   99.61 & 97.04  /   90.83 & 94.68  /   87.04 & 99.91  /   99.70\\
\textbf{Elastic Cos+} &   99.89  /   99.83 & 98.62  /   91.11 & 95.35  /   92.22 & 99.89  /   99.66 & 98.02  /   88.84 & 96.20  /   82.22 & 99.89  /   99.66\\
\hline  
\hline 
\end{tabular}
}
\label{tab:verid}
\end{table*} 

\subsection{Face image quality assessment}
 We further evaluate the utility of synthetically altered images by studying the EDC to verify the image quality usable within FRS. EDC curve shows the false non-match error rate on the y-axis achieved when discarding a certain percentage of low-quality images (x-axis). Based on the predicted quality values, these discarded images have the lowest predicted quality, and the error rate is calculated on the remaining images. EDC curve indicates good quality estimation when the verification error decreases consistently when the ratio of unconsidered images increases. We employ the Arcface FRS described to measure the errors vs discard characteristics. The verification error is reported regarding the false non-match rate (FNMR) at fixed false match rates (FMR). The FMR is reported at 0.05\% FMR threshold. 
 
 This analysis establishes the use of ethnicity-altered images for biometric applications as they follow the trends of real data. As noted from Fig ~\ref{fig:edc}, generic trends of the EDC are followed for real data and ethnicity-altered images across different architectures used for ethnicity alteration. Examining the CycleGAN EDC plots reveals that MagFace and CRFIQA yield quality scores that reduce the FNMR, whereas SERFIQ appears to elevate the FNMR. In contrast, StarGAN EDC plots indicate that MagFace and CRFIQA outperform SERFIQ when paired with ArcFace. 

 In terms of ArcFace FRS models, MagFace emerges as the top performer overall, closely followed by CRFIQA. Moreover, ethnicity-altered images by CycleGAN and StarGAN, which leverage image-to-image translation models, exhibit fewer artifacts compared to FGAN, which relies on manifold learning. Notably, all FIQA methods demonstrate enhanced performance on Asian subjects generated by CycleGAN, StarGAN, and FGAN. The results of CRFIQ and MagFace remain consistent across Asian, Black, and Indian subjects. However, a pronounced bias towards Black subjects is evident across all FRS-FIQA model combinations.

\subsection{Cross Ethnicity Face Verification}
Finally, we assess face verification performance in cross-ethnicity protocol for images generated using CycleGAN, StarGAN, and FGAN using four distinct FRS: FaceNet \cite{FaceNet2015}, PFE \cite{PFE2019}, and two variants of ElasticFace \cite{Boutros_2022_CVPR}. It is anticipated that a high FRS performance is obtained when a higher degree of identity is preserved. For this analysis, the embeddings are extracted from the pre-trained FRS models and the distance between them is calculated using the cosine distance.
	
We thoroughly evaluated the FRS on the RFW test datasets by creating mated and non-mated image pairs. To create mated pairs, we utilized the generated images of each target domain and the input images of the source domain ethnicity for individual subjects. Furthermore, we randomly generated cross-subject imposter pairs for comparison. The results of SOTA FRS evaluation are presented in Table~\ref{tab:verid} and illustrate the system's performance at the FNMR with an FMR of $0.1\%$ and $0.01\%$.

It can be noticed that CycleGAN and StarGAN verification rates are lower than FGAN due to the latter generating better ethnicity-altered images (as seen in Figure~\ref{fig:treeganasian}, Figure~\ref{fig:treeganblack}, and Figure~\ref{fig:treeganindian}) and the distribution of ITA angles between altered and real images overlaps, as illustrated in Figure~\ref{fig:skintoneita}. Also, altered Asian and Black images using CycleGAN exhibit minor differences in verification accuracy compared to real and cross-ethnicity altered images. Additionally, altered Indian ethnicity images using StarGAN demonstrate better verification accuracy than those altered using CycleGAN, suggesting that these models are more effective for generating cross-altered ethnicity synthetic datasets. On an average, all FRS perform better for Indian ethnicity than Asian and Black ethnicity across real and altered images. 
Indeed, it is notable that the accuracy of cross-ethnicity face verification increases when utilizing newer models such as Elastic-Arc+ and Elastic-Cos+, as opposed to older models such as FaceNet and PFE, both for mated and non-mated score distributions on both real and altered images. 

\section{Ethnicity Alteration Challenges}\label{sec:challenges}
\begin{figure}[htp]
    \centering
    \includegraphics[width=9.0cm]{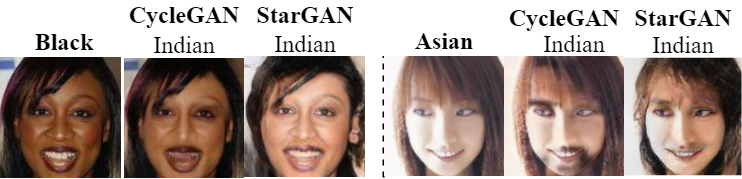}
    \caption{Ethnicity alteration challenging examples, e.g., the occluded forehead woman’s identity is transferred to man.}
    \label{fig:treeganlimitation}
    \vspace*{-0.3cm}
\end{figure}

Ethnicity alteration can face challenges as depicted in Figure \ref{fig:treeganlimitation} occasionally. Two sample cases where ethnicity alterations using CycleGAN and StarGAN fail is presented in Figure \ref{fig:treeganlimitation}. Specifically, it showcases instances in which a woman's identity is erroneously transferred to a man. Such potential pitfalls of ethnicity alteration techniques underscore the need for further refinement and improvement in the development of these models, ensuring more reliable and accurate results.

   
\section{Discussion}\label{sec:discussion} 
Based on the analysis conducted in this work, we observe the following for the research questions formulated in Section \ref{sec:introduction}:

\begin{itemize}
    \item \textbf{Q1}. Does synthetic ethnicity alteration and skin tone modification using Generative Adversarial Networks (GANs) find use in FRS for representing a subject in different ethnicities?
    
    \textbf{A}. Our analysis reveals that the accuracy of verifying altered images of Asian, Black, and Indian ethnicities using CycleGAN is comparable to that of real images of different ethnicities suggesting the ethnicity alteration and skin tone modification can be used for biometric applications. Further, CycleGAN is more effective at retaining identity information than StarGAN and FGAN. Moreover, ElasticArc+ and ElasticCos+ exhibit superior performance on synthetically altered images compared to the other two legacy FRS. 
    
    \item \textbf{Q2}. Does synthetic ethnicity alteration and skin tone modification of a subject's facial image resemble ITA of real faces?

    \textbf{A}. Based on the extensive experiments presented in Figure \ref{fig:skintoneita}, it is evident that ITA remains relatively stable for face images with altered ethnicity utilizing StarGAN and FGAN, similar to real face images. This observation suggests synthetic alteration approaches effectively transfer skin tones to different ethnicities. However, a deviation from this stability is noticeable for black ethnicity when generated using CycleGAN suggesting further investigations. CycleGAN introduces variations in ITA when changing ethnicities, potentially affecting skin tone representation accuracy for certain ethnicities compared with real faces.
    
    \item \textbf{Q3}. Do FIQA and FRS measures for synthetic ethnicity-altered image datasets indicate dependable performance, similar to real datasets of different ethnicities? 
    \textbf{A}. Our analysis on synthetically generated cross-ethnicity data through image-to-image translation techniques suggests that both the FIQA and FRS provide performance similar to real datasets. Leveraging these measures can facilitate the creation of comprehensive training datasets while maintaining privacy using image-to-image translation approaches.
    
\end{itemize}
		\section{Conclusion and Future work}\label{sec:conclusion}
		The feasibility of modifying ethnicity and skin tone biometric applications was presented in this work. A new dataset featuring an equal representation of three ethnicities: Asian, Black, and Indian was constructed to generate images of different ethnicity representation. Generative adversarial network-based image-to-image translation models and manifold learning based models were used to generate images of altered ethnicity. With a detailed analysis on skin-tone conformance across ethnicity, FIQA coherence similar to real datasets and FRS performance similar to real data of different ethnicity, this work has shown the possibility of altering ethnicity for biometric applications. The study points to possible use of synthetic data to mitigate bias within the FRS. Synthetic data enables the representation of diverse demographic groups while maintaining consistent image quality across ethnicities and realistic intraclass variation as exemplified in this work. Future research could focus on extending image-to-image translation with recent diffusion models for creating multiple ethnicity representation and use it for making FRS more robust and bias-free.


\balance

\bibliographystyle{IEEEtran}
\bibliography{mybibfile}

 	\end{document}